\documentclass{article}
\usepackage{graphicx}
\usepackage[final]{corl_2026}
\usepackage{float}
\usepackage{gensymb}
\usepackage{amssymb}
\usepackage{amsmath}
\usepackage{dsfont}
\usepackage{booktabs}
\usepackage{array}
\usepackage{wrapfig}
\usepackage{algorithm}
\usepackage{algpseudocode}
\usepackage{multirow}
\usepackage{bm}
\usepackage{enumitem}

\newcommand{\bfitnum}[1]{\bm{\mathit{#1}}}

\title{Transformer-Based Trajectory Prescription for sEMG-Optimizing Robot-Mediated Rehabilitation}

\author{
  Gabriel Parra$^{1,2}$, Kimia Khoshnami$^1$, Daniel S. Brown$^2$, and Laura A. Hallock$^{1,2}$\\
  $^1$Department of Mechanical Engineering and $^2$Kahlert School of Computing, University of Utah\\
  \texttt{\{u1289824,kimia.khoshnami,daniel.s.brown,laura.hallock\}@utah.edu} %
}

\begin{document}
\maketitle

\begin{abstract}

Robot-mediated rehabilitation has the potential to deliver high-intensity, individualized therapy, but existing systems (which have not consistently improved outcomes) typically prescribe kinematic trajectories rather than explicitly targeting the neuromotor activations thought to support recovery. Prescribing motions that elicit desired activity is challenging because muscle activity is hysteretic and stochastic, noninvasive measurement methods (e.g., surface electromyography, or sEMG) are noisy and sensor-location-dependent, and  interventional datasets are costly to collect, inhibiting model development. To address this gap, and ultimately enable activation-optimizing rehabilitation robot behavior, we present a personalizable transformer-based model mapping individuals' muscle activations (as measured via sEMG) to prescribed contact force trajectories when interacting with an end-effector rehabilitation robot. The model uses a transformer encoder--decoder leveraging rotary positional embeddings (RoPE) to generate a latent representation of trajectory-relevant and user-specific neuromotor factors, and augments trajectory prediction with flow matching to produce a distribution over plausible prescriptions biased toward the training distribution. Using data from 16 healthy participants completing 15 isometric trajectory tracking tasks, we evaluate generalization across held-out tasks and users, as well as the feasibility of user-specific fine-tuning within a therapy-session timescale. The model achieves mean trajectory prediction errors of $<$8\% on held out tasks and users, with effectively instantaneous deterministic inference and sub-second sampling of trajectory distributions. User-specific fine-tuning further reduces error using only minutes of additional data collection and training. An initial test using the model during live rehabilitation tasks demonstrates that, on average, participants reach the target neuromotor activations with just $15\%$ error accumulated throughout the task. These results provide a proof of concept for data-driven, activation-targeted trajectory prescription in robot-mediated rehabilitation and establish a foundation for future validation in clinical populations. To enable further development by scientific and user communities,
data and code have been open-sourced
as part of the OpenRobotRehab project on SimTK ({\tt \href{https://simtk.org/projects/openrobotrehab}{simtk.org/projects/openrobotrehab}}).

\end{abstract}

\section{Introduction}
\vspace{-0.5em}

Robot-mediated therapy is a promising 
treatment for 
neuromotor injury (e.g., stroke), but existing systems (which largely rely on impedance-style error correction to desired kinematic trajectories) have not yet demonstrated meaningful improvement in rehabilitation outcomes over standard of care~\cite{park2024effects}. Neuroscientists posit that improving such outcomes requires systems to reinforce healthy neuromotor pathways --- i.e., induce healthy patterns of underlying muscle activation, not just guide the user through desired kinematic motions~\cite{kleim2008principles, maier2019principles}. 

An ideal robotic rehabilitation system (of the type illustrated in Figure~\ref{fig:platform}) would allow a new patient to arrive at a clinic, interact with the robot briefly to calibrate the system, then perform a set of therapy tasks that induce healthy activations for that individual. (Note that what constitutes ``healthy'' motion is an active area of investigation~\cite{ting2012review, anand2026characterizinghealthypoststroke}; in this work, we consider this to be a therapist- or expert-specified time series of muscle activations.) However, there do not yet exist models relating prescribed therapy activities (i.e., trajectories) to neuromotor activations of a given individual: even the healthy human neuromotor system exhibits complex (hysteretic and probabilistic) behavior, noninvasive measurement methods (e.g., surface electromyography, or sEMG) are noisy and dependent on sensor location, and large-scale data of patient populations are prohibitively expensive to collect, inhibiting both physics-based and data-driven model development.

In this work, we leverage the robust modeling capabilities of transformers~\cite{vaswani2023attentionneed} and flow matching~\cite{lipman2024flowmatchingguidecode} to construct a trajectory prescription model with the following properties:
(1) given a target sEMG-measured time series of muscle activations, predict the required target trajectory to induce these activations; 
(2) require minimal per-user calibration accomplishable within a single therapy session (i.e., taking only a few minutes);
(3) once calibrated, predict trajectories in a sufficiently short time (i.e., a few seconds maximum) to enable a therapeutically meaningful number of repetitions and distinct motions within a session; and
(4) provide well-calibrated uncertainty modeling to flag out-of-distribution target muscle activations, supporting decisions about when prescribed trajectories should (and should not) be trusted.
Specifically, we present the following contributions: 
\begin{itemize}[leftmargin=1.2em, nosep]
    \item 
    Construction and training of a flow-matching-augmented, transformer-based trajectory prescription model mapping desired sEMG-measured muscle activations of the arm to the associated end-effector (hand) force target trajectory, quickly fine-tunable to new users.
\item 
Demonstration in an exemplar cohort of 16 healthy users performing 15 trajectory tracking tasks that our model achieves an average of $<$8\% trajectory prediction error on held-out users and tasks (with the bulk of that error attributable to the conservative, training-distribution-seeking behavior of the flow matching algorithm), and that user-specific calibration can substantially reduce that error within a few minutes' data collection and training with quantifiable time--performance tradeoffs.
\item 
Deployment of the model during live rehabilitation tasks for 5 healthy users, demonstrating an average task-accumulated error between model-prescribed and sEMG-measured user neuromotor activations of $15\%$.
\item 
The open-source release of all data and modeling code as part of the OpenRobotRehab project on SimTK ({\tt \href{https://simtk.org/projects/openrobotrehab}{simtk.org/projects/openrobotrehab}}), enabling both additional insights on neuromotor behavior and further model augmentation and population-specific refinement by the wider rehabilitation and machine learning communities.
\end{itemize}

We present this work as a first-of-its-kind, proof-of-concept system: our preliminary cohort establishes the feasibility of activation-targeted trajectory prescription, laying the foundation for expansions toward numerous clinical applications (e.g., post-injury muscle-specific strength training, post-stroke muscle coordination training).
\vspace{-0.5em}

\section{Related Work}\label{sec:related}
\vspace{-0.5em}

Principles of experience-dependent plasticity and motor learning suggest that effective neuromotor therapy should reinforce healthy patterns of neuromotor engagement --- i.e., time series muscle activations ---
rather than merely guiding the limb through desired motions~\cite{kleim2008principles, maier2019principles}. While even healthy motor behavior is heterogeneous~\cite{anand2026characterizinghealthypoststroke}, properties of desirable and undesirable activation patterns can be characterized across a wide range of therapy applications: in stroke, for example, a therapeutic target is often to move outside of abnormal muscle synergies and avoid excessive co-contraction~\cite{dipietro2007changing}, and in sports medicine, increasing activation of under-recruited muscles and improving neuromotor coordination can be used to avoid injury~\cite{griffin2003neuromuscular}.
At the same time, clinical rehabilitation rarely incorporates these insights directly, and remains largely empirical: in practice, therapists select movements, assistance, resistance, and perturbations based on experience, without a systematic way to choose tasks that induce a desired pattern of neuromotor activation.

Robot-mediated rehabilitation, alongside neuromechanical sensing of muscle activations, kinematics, and contact forces, is a promising paradigm to systematize the prescription of therapy tasks to best induce such activations.
In the literature, if not in practice, sEMG biofeedback has long been used to expose muscle activity to patients and clinicians, supporting efforts to increase weak muscle recruitment, reduce inappropriate activation, or encourage task-oriented retraining~\cite{huang2006recent, giggins2013biofeedback}. 
Recent work (e.g., Lee et al.~\cite{lee2025muscle}) has begun to connect such muscle-level targets to prescribed task-level actions, demonstrating significant therapeutic effects in pilot cohorts that support the feasibility of such an approach. At the same time, these systems typically remain reliant on 
explicit pathology- or task-specific models, inhibiting translation to new populations and motions. Those that aim to model task-agnostic behavior at the neuromuscular level generally employ neuromusculoskeletal simulators (e.g., OpenSim~\cite{delp2007opensim}, MyoSuite~\cite{caggiano2022myosuite}) that are expensive to calibrate and may not generalize across individuals, sensor placements, impairments, or contact conditions.

The heterogeneity and stochastic nature of neuromotor behavior makes machine learning a natural approach to close this modeling gap, and ML-based models have been widely applied to sEMG for gesture recognition, prosthetic and orthotic control, intent inference, torque estimation, and human--robot interaction~\cite{tamilvanan2025comprehensive, zheng2022surface, sid2024comprehensive}, though never (to our knowledge) to the problem of target trajectory generation during robot-mediated rehabilitation. 
To construct such a model, we employ a transformer encoder--decoder architecture~\cite{vaswani2023attentionneed} (detailed in Section~\ref{sec:model}), motivated by its strong performance in related sEMG inference tasks (e.g.,~\cite{xu2024chatemg, wang2024transformer, zhang2023lst}). This architectural choice is well-suited to sEMG, which is a multichannel temporal signal whose interpretation depends on activation history, co-activation patterns, and task phase. Self-attention enables the model to capture dependencies across an entire time window without requiring hand-engineered temporal features or strictly recurrent state updates~\cite{vaswani2023attentionneed}, and transformer-based representation learning has proven effective for multivariate time-series modeling~\cite{zerveas2021transformer}.
We use rotary positional embeddings (RoPE)~\cite{su2023roformerenhancedtransformerrotary} to preserve temporal ordering while allowing attention to depend explicitly on relative time offsets, which is important when mapping activation trajectories rather than isolated samples. In designing our architecture, we also acknowledge that the desired inverse map is not a deterministic function of observed sEMG alone: neuromechanical redundancy, user-specific pursuit strategies, electrode placement, fatigue, and impairment introduce unobserved factors that can change the relationship between a displayed trajectory and the resulting muscle activity. This motivates a compact latent-variable representation that separates target-associated trajectory information from latent neuromotor behavior, following the broader use of continuous latent representations for scalable inference in partially observed systems~\cite{kingma2022autoencodingvariationalbayes} and aligning with neuromechanical understanding of redundant muscle activation and individual variation~\cite{ting2012review, stanev2019modeling}. Finally, because multiple force or velocity trajectories may plausibly induce similar desired activation patterns, we model a conditional distribution over latent trajectory codes rather than only a point estimate. Flow matching provides an efficient simulation-free objective for learning continuous generative flows~\cite{lipman2024flowmatchingguidecode}, including conditional variants suitable for sampling from learned conditional distributions~\cite{tong2024improvinggeneralizingflowbasedgenerative}; in our setting, these samples can be decoded into a distribution over prescribed trajectories, providing an uncertainty estimate that can inform when additional user-specific calibration may be needed. Flow matching also tends to bias prescribed trajectories toward the distribution of the training data set --- a useful (though not guaranteed) property, as out-of-distribution trajectories may be unsafe to execute on a robot. Section~\ref{sec:model} describes how each of these design choices is realized in our architecture.

\vspace{-0.5em}
\section{System Design}
\vspace{-0.5em}

This section presents our trajectory prescription model (Section~\ref{sec:model}) for a robot-mediated rehabilitation platform (Section~\ref{sec:platform}), trained and tested using data collected from 16 participants (Section~\ref{sec:data}).

\begin{figure}[t]
    \centering
    \includegraphics[width=0.8\linewidth]{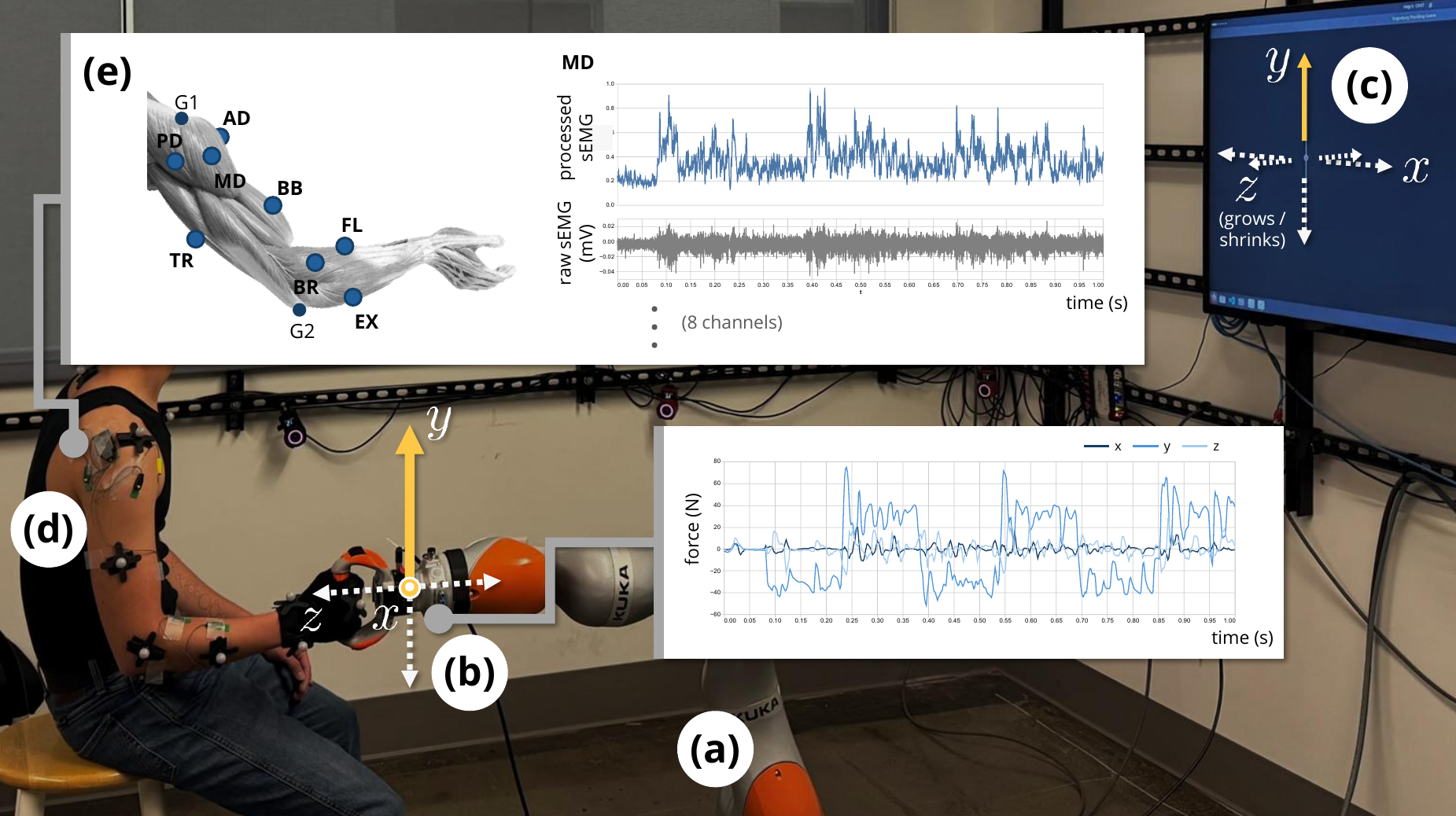}
    \caption{\small
        End-effector robot-mediated rehabilitation platform. Users exert forces on a LBR iiwa 14 R820 7-degree-of-freedom cobot (KUKA AG, Augsburg, Germany; a) via a custom handle attached to a SensONE 6-axis force torque sensor (Bota Systems AG, Zürich, Switzerland; b). Users' forces along each positional axis are mapped to the corresponding $x/y/z$ velocity of an on-screen avatar ball (c), where the $z$ axis into and out of the screen is represented by a decrease and increase in avatar size, respectively. Users are instructed to follow a target ball along a displayed trajectory. Force and game trajectory data (of both avatar and target) are collected at 1000~Hz and 60~Hz, respectively. During this process, two Trigno Quattro 4-channel sensor motes (d), controlled through a Trigno Base Station (Delsys, Inc., Natick, MA, USA) collect surface electromyography (sEMG) data at a rate of 2.22~kHz from 8 upper limb muscles (e). Electrode placements are as follows: anterior deltoid (AD), middle deltoid (MD), posterior deltoid (PD), and biceps brachii (BB), grounded at the acromion (G1); and triceps brachii (long head, TR), brachioradialis (BR), wrist flexors (FL), and wrist extensors (EX), grounded at the olecranon (G2). All data are collected and synchronized via Redis (MongoDB, Inc., New York, NY, USA). Full implementation details are included with code release. 
        \vspace{-1.5em}
    }
    \label{fig:platform}
\end{figure}

\subsection{Robot-Mediated Rehabilitation Platform}\label{sec:platform}
\vspace{-0.25em}

Figure~\ref{fig:platform} illustrates our end-effector robot-mediated rehabilitation platform, an extension of Anand et al.'s OpenRobotRehab platform~\cite{anand2025icorr}.

\subsection{Data Collection}\label{sec:data}
\vspace{-0.25em}

Data were collected from 16 healthy participants (P01--P16; 5 female, 11 male; age: $28.4 \pm 7.0$ (mean $\pm$ standard deviation, min 19, max 47); 5 left-handed, 11 right-handed), none of whom had prior experience performing this study's tasks. 
Three additional participants (P17, P18, P19) did not complete initial data collection but participated in preliminary closed-loop trajectory prescription tests discussed in Section~\ref{sec:baseline}.\footnote{The study protocol employed in this work was approved by the University of Utah Institutional Review Board under Protocol IRB\_00183389, and written informed consent was obtained from each study participant. Full demographic data are included in Appendix~\ref{app:pardem}.}

Each participant executed 15 isometric force trajectories (see Figure~\ref{fig:platform} (d)), all with their (sEMG-instrumented) right arm regardless of handedness, designed to traverse the user's $x/y/z$ force workspace, in the order enumerated in Table~\ref{tab:traj}\footnote{Several minor deviations from this protocol are documented in Appendix~\ref{app:prodev}.}. Tasks included single axis motions (T01--T12)
and stochastically generated trajectories surrounding the $x$ and $y$ axes generated via a smoothed Brownian motion (T14--T15).

\begin{table}[t]
\footnotesize
\def\arraystretch{1.1}%
\caption{Trajectory Tracking Tasks}
    \label{tab:traj}
    \renewcommand{\arraystretch}{1.3}
\begin{tabular}
{>{\raggedright\arraybackslash}p{0.04\textwidth} | >{\raggedright\arraybackslash}p{0.13\textwidth} | >{\raggedright\arraybackslash}p{0.075\textwidth} | >{\raggedright\arraybackslash}p{0.2\textwidth} | >{\raggedright\arraybackslash}p{0.05\textwidth} | >{\raggedright\arraybackslash}p{0.05\textwidth} | >{\raggedright\arraybackslash}p{0.095\textwidth} | >{\raggedright\arraybackslash}p{0.04\textwidth} | >{\raggedright\arraybackslash}p{0.03\textwidth}} 
\hline
Task No. & Task & Other axes locked? & Force required for unit velocity $x/y/z$ motion (N) & Repe-ti-tions & Dura-tion (s) & Trajectory displayed? & Trai-ning set? & Test set? \\ \hline
T01 & $x$-axis & No & $15/15/36$ & 3 & 60 & Yes & $\checkmark$ & \\
T02 & $y$-axis & No & $15/15/36 $ & 3 & 60 & Yes & $\checkmark$ & \\
T03 & $z$-axis & No & $15/15/36$ & 3 & 60 & Yes & $\checkmark$ & \\
T04 & $x$-axis & Yes & $15/-/-$ & 3 & 60 & Yes & $\checkmark$ & \\
T05 & $y$-axis & Yes & $-/15/-$ & 3 & 60 & Yes & $\checkmark$ & \\
T06 & $z$-axis & Yes & $-/-/36 $ & 3 & 60 & Yes & $\checkmark$ & \\
T07 & $x$-axis & No & $30/30/72$ & 3 & 60 & Yes & $\checkmark$ & \\
T08 & $y$-axis & No & $30/30/72$ & 3 & 60 & Yes & $\checkmark$ & \\
T09 & $z$-axis & No & $30/30/72$ & 3 & 60 & Yes & $\checkmark$ & \\
T10 & $x$-axis & Yes & $30/-/-$ & 3 & 60 & Yes & $\checkmark$ & \\
T11 & $y$-axis & Yes & $-/30/-$ & 3 & 60 & Yes & $\checkmark$ & \\
T12 & $z$-axis & Yes & $-/-/72$ & 3 & 60 & Yes & $\checkmark$ & \\
T14 & stochastic-$x$ & No & $20/20/48$ & N/A & 20 & Yes & & $\checkmark$ \\
T15 & stochastic-$y$ & No & $20/20/48$ & N/A & 20 & Yes & & $\checkmark$ \\ \hline
\end{tabular}
\end{table}

Trajectories were performed at 2 distinct force levels and both with and without non-targeted axes ``locked'' for single axis tasks (e.g., T01 and T04 represented the same $x$-axis traversal task, but for T01, the avatar moved in all dimensions such that the user must maintain $y$ and $z$ force values at 0 for accurate task completion, while for T04, the avatar was locked to the $x$-axis such that these $y$ and $z$ force values could vary freely without impacting task performance).\footnote{Locked tasks facilitated smooth, natural motion, but admitted arbitrary ``null-space forces,'' known to result in varied behavior across users~\cite{anand2026characterizinghealthypoststroke}; unlocked tasks enforced that all participants attempted to execute the exact same force trajectory, but empirically resulted in varied perturbations as users drifted from, then corrected to, each axis. Motivation for inclusion of both trajectory types, and analysis of the model's performance thereon, can be found in Appendix~\ref{app:lock}; in our primary analyses, these tasks are treated equivalently, with locked axes corresponding to a target value of 0 for all time points.} Visuals for each trajectory are available in Appendix~\ref{app:traj}.

\subsection{Trajectory Prescription Model}\label{sec:model}

Recall that we aim to prescribe a target trajectory to display to a given user to induce a specified pattern of sEMG-measured muscle activations.
Mathematically, this means that we seek to identify some function mapping desired sEMG muscle activation time series $\mathbf{a} = \{\mathbf{a}_{t} \in \mathbb{R}^8\}_{t=1}^T$ over $T$ time steps
to target $x/y/z$ velocity trajectory $\mathbf{v}_\mathrm{target} = \{\mathbf{v}_{\mathrm{target},t} \in \mathbb{R}^3\}_{t=1}^T$  that will induce these activations. We know, however, that target trajectory is not determined by sEMG content alone: even if a user completes a trajectory perfectly, neuromechanical redundancy~\cite{stanev2019modeling} means that a user may engage a continuum of possible muscles, and in practice, users will complete the tracking task with varying skill levels, and employ varying pursuit strategies~\cite{anand2026characterizinghealthypoststroke}. These strategies vary by user, as do noise characteristics of the sEMG signal due to variations in sensor placement, attenuation due to adipose tissue, and fatigue. We therefore model the activation--target relationship $f$ as
$$\mathbf{a}=f(\mathbf{v}_\mathrm{target}, U)$$
for some unknown latent variable $U$ encompassing this variation in neuromechanical behavior. 
Our task of trajectory prescription can then be written as: model some function $g \approx f^{-1}$, such that we can prescribe some trajectory $\mathbf{\hat v}_\mathrm{target}$ via
$$\mathbf{\hat v}_\mathrm{target}=g(\mathbf{a})=g(f(\mathbf{v}_\mathrm{target}, U))$$
given some desired activation $\mathbf{a}$.

In addition to accounting for this unknown latent variable $U$, the function $g$ must be sufficiently expressive to account for the time-dependent nature of the sEMG signal. (An instantaneous sEMG sample is meaningless alone, and a function of both prior and future sEMG data.) Practically, any model for $g$ must also admit rapid retraining to accommodate out-of-distribution users, and, ideally, provide a measure of uncertainty on trajectory predictions to inform decision making on when model updates are required.

\begin{wrapfigure}{r}{0.6\textwidth}
    \centering
    \includegraphics[width=1.0\linewidth]{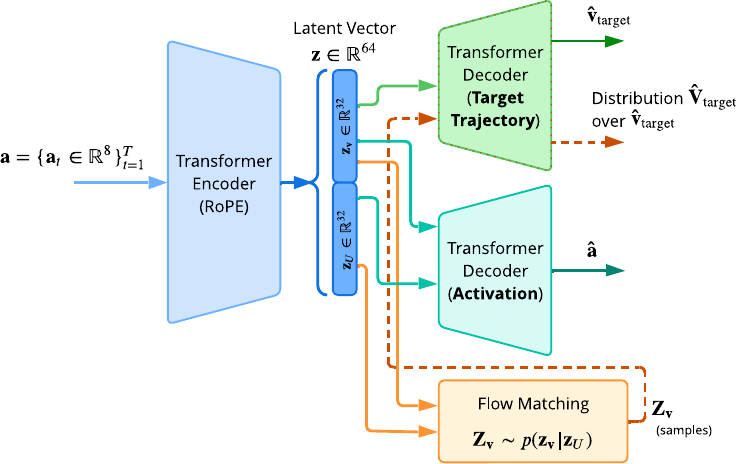}
    \caption{\small Architecture of trajectory prescription model. Latent representation $\mathbf{z}$ of both target-associated information ($\mathbf{z}_{\mathbf{v}}$) and unknown neuromotor behavior ($\mathbf{z}_U$) is generated using a RoPE~\cite{su2023roformerenhancedtransformerrotary} transformer encoder (\textit{blue} flow), then decoded separately to generate prescribed trajectory $\mathbf{\hat v}_\mathrm{target}$ (\textit{green} flow) and estimated activation $\mathbf{\hat a}$ (\textit{turquoise} flow). A flow-matching diffusion model~\cite{lipman2024flowmatchingguidecode} is also trained to generate a conditional distribution of $\mathbf{z}_{\mathbf{v}}$ given $\mathbf{z}_U$ whose samples can be passed through the target trajectory decoder to generate a distribution over  $\mathbf{\hat v}_\mathrm{target}$ (\textit{orange} flow).}
    \vspace{-1em}
    \label{fig:arch}
\end{wrapfigure}

Consistent with these design constraints, we present the following transformer--autoencoder architecture model for $g$, illustrated in Figure~\ref{fig:arch}. Activations $\mathbf{a}$ 
are encoded via a Rotary Position Embedding (RoPE)~\cite{su2023roformerenhancedtransformerrotary} transformer into a latent representation $\mathbf{z} \in \mathbb{R}^{64}$ consisting of concatenated vectors $\mathbf{z}_{\mathbf{v}} \in \mathbb{R}^{32}$ and $\mathbf{z}_U \in \mathbb{R}^{32}$ compressing information corresponding to the target trajectory and unknown neuromechanical behavior, respectively. This partition mirrors the structure of $f$ above (and the motivation in Section~\ref{sec:related}): the activations a user generates are a function of both the prescribed trajectory and that user's specific motor strategy, so we allocate one latent to each factor. 
Separate transformer decoders then reconstruct the prescribed target trajectory $\mathbf{\hat v}_\mathrm{target}$ from $\mathbf{z}_{\mathbf{v}}$ (the primary model output) and estimated associated activations $\mathbf{\hat a}$ from $\mathbf{z}_{\mathbf{v}}$ and $\mathbf{z}_U$ (autoencoder only used in training). To model uncertainty over $\mathbf{\hat v}_\mathrm{target}$, a flow-matching diffusion model~\cite{lipman2024flowmatchingguidecode} is also trained 
to generate conditional probability distribution $p(\mathbf{z}_{\mathbf{v}}|\mathbf{z}_U)$. An arbitrary number of samples $\mathbf{Z}_{\mathbf{v}}$ can then be generated from this distribution and passed through the target trajectory decoder to generate a distribution $\mathbf{\hat V}_\mathrm{target}$ over $\mathbf{\hat v}_\mathrm{target}$. In addition to providing a measure of prediction uncertainty, this flow matching formulation tends to bias samples of $\mathbf{\hat V}_\mathrm{target}$ toward the distribution of the training set, with uncertainty generally rising for inputs sparsely represented or absent in the training data: while this may curtail generalization outside this distribution, we consider it a useful property for our application, as prescribed trajectories far outside of distribution may be unsafe to execute. We emphasize, however, that this behavior is not a theoretical safety guarantee. Instead, we envision flow matching uncertainty as a decision-support tool within the therapy workflow: if the model exhibits high uncertainty for a requested set of target muscle activations, the system (or supervising therapist) should defer targeting those activations until more information about the user has been gathered, e.g., via the fine-tuning procedure of Section~\ref{sec:finetuning}.

The system is then jointly trained to minimize loss function 
\begin{equation}\label{eq:loss}
    \mathcal{L}=\lambda_1 \left\lVert\mathbf{v}_\mathrm{target} - \mathbf{\hat v}_\mathrm{target}\right\rVert_2^2+\lambda_2 \left\lVert\mathbf{a} - \mathbf{\hat a}\right\rVert_2^2+\lambda_3 \left\lVert\mathbf{z}_{\mathbf{v}} - \mathbf{Z}_{\mathbf{v}}\right\rVert_2^2
\end{equation}
such that primary trajectory prediction $\mathbf{\hat v}_\mathrm{target}$ is regularized by activation decoder and flow-matching terms to encourage a structured latent representation and uncertainty-aware trajectory prescription, respectively.

\vspace{-0.5em}
\section{Experiments}
\vspace{-0.5em}

Motivated by our intended use case --- a new user briefly interacts with the platform, after which the system prescribes trajectories based on expert-specified desired sEMG patterns --- we present the following evaluations of model performance in terms of both prescribed trajectory accuracy and required training time.

\subsection{Baseline Model Training \& Evaluation}\label{sec:baseline}

We first characterize model performance absent any user-specific retraining. 

\textbf{Training and evaluation splits.}
The baseline model was trained using data from participants P01--P10, tasks T01--T12, as noted in Table~\ref{tab:traj}, and evaluated on unseen tasks for known users (P01--P10, T14--T15), known tasks for unseen users (P11--P16, T01--T12), and unseen tasks for unseen users (P11--P16, T14--T15). Note that this division results in a training set that contains highly stereotyped, single-axis motions, and unseen-task test sets that are stochastic in nature and thus contain trajectory regions both in and out of distribution of the test set, allowing exploration of performance across these conditions.
The final data set contained 6480 training windows and 4562 test windows.

\textbf{Model training.}
Our (Figure~\ref{fig:arch}) model was trained to minimize the loss function in Equation~\ref{eq:loss}. 
Our implementation used PyTorch with AdamW using a learning rate of 0.001, batch size of 64, and was trained for 100~epochs.
Each training example consisted of a 10~s window (generated with a 1~s stride across each 60~s trial) of processed sEMG \footnote{See Appendix~\ref{app:emgproc} for additional details on sEMG processing.} activation $\mathbf{a}$ paired with corresponding target trajectory $\mathbf{v}_{\mathrm{target}}$. 
Windows were shuffled and dropped (dropout $p$=0.1) during training. Flow matching time steps were sampled from a uniform distribution, with base noise sampled from the standard normal distribution and 10~samples drawn per training example.
Hyperparameter selection is documented in Appendix~\ref{app:hyper}.
Training was performed on a NVIDIA RTX 4070 with 12~GB of VRAM. A full baseline training run required $\approx$1.5~h. At inference time, generating one prescribed trajectory was effectively instantaneous (5~ms), and generating 250 flow matching samples required $<$1~s.

\textbf{Trajectory prediction performance metrics.}
We quantify model performance as mean squared prediction error for both our deterministic transformer-decoded target trajectory ($\mathrm{MSE}_\mathrm{TD}$) and our flow matching distribution model ($\mathrm{MSE}_\mathrm{FM}$), defined as
\begin{align*}
\mathrm{MSE}_\mathrm{TD} = \frac{1}{n_{\mathrm{trials}} n_\mathrm{windows}}\sum_{i=1}^{n_{\mathrm{trials}}}\sum_{j=1}^{n_{\mathrm{windows}}}\left\lVert \mathbf{v}_\mathrm{target} - \mathbf{\hat{v}}_{\mathrm{target}}\right\rVert_{2_{i,j}}^2 \\
\mathrm{MSE}_\mathrm{FM} = \frac{1}{n_{\mathrm{trials}} n_\mathrm{windows}}\sum_{i=1}^{n_{\mathrm{trials}}}\sum_{j=1}^{n_{\mathrm{windows}}}\left\lVert \mathbf{v}_\mathrm{target} - \mathbb{E}[\hat{\mathbf{V}}_\mathrm{target}]\right\rVert_{2_{i,j}}^2
\end{align*}
where $\mathbb{E}[\hat{\mathbf{V}}_\mathrm{target}]$ is computed as the sample mean across 25 trajectories per activation window, each decoded from an independent flow matching sample of $\mathbf{Z}_{\mathbf{v}}$. (See code release for implementation details.) We selected 25 samples per window to generate a variety of candidate trajectories while remaining within our GPU's memory limits, enabling rigorous evaluation across all test windows.

\begin{figure}[t]
    \centering
    \includegraphics[width=1.0\linewidth]{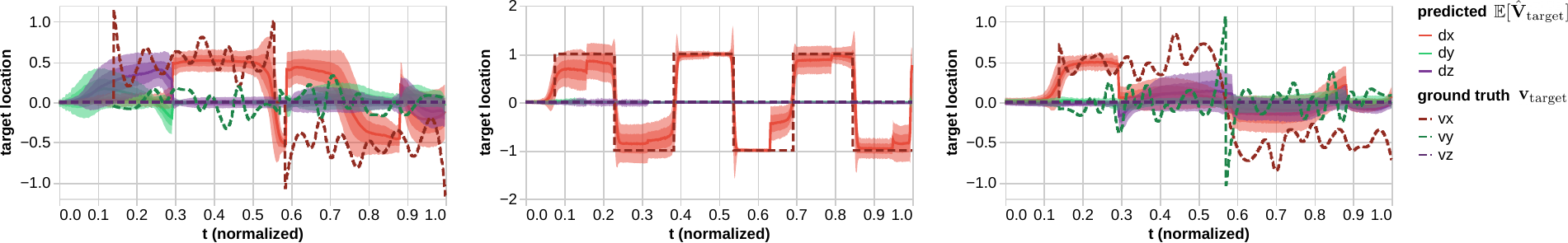}
    \caption{\small
    Exemplar trajectory prediction accuracy $\hat{\mathbf{V}}_\mathrm{target}$ distributions for unseen tasks and unseen users.
    \textit{Left}: Known participant (P09) performing unknown task (T14). For regions within the distribution of tasks observed during training ($0.3<t<0.6$), trajectory mean is accurately predicted. For regions outside this distribution ($t<0.3$, $t>0.9$), flow matching tends to pull predicted trajectories toward in-distribution regions, which are more likely to be safe. Some regions ($0.6<t<0.9$) exhibit a mix of this behavior across axes.
    \textit{Center}: Unseen participant (P11) performing known task (T07). Trajectory predictions are accurate aside from artifacts (e.g., at $t\approx 0.15$) due to the length of windows that were seen by the transformer at training time. \textit{Right}: Unseen participant (P16) performing unseen task (T14). Prediction is qualitatively comparable to unseen tasks performed by known users, with slightly higher overall error rates.}
    \label{fig:baseline}
    \vspace{1em}
    \includegraphics[width=0.3\linewidth]{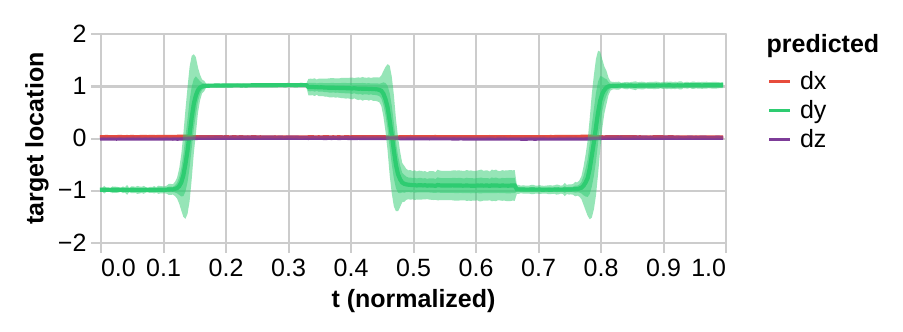}\includegraphics[width=0.7\linewidth]{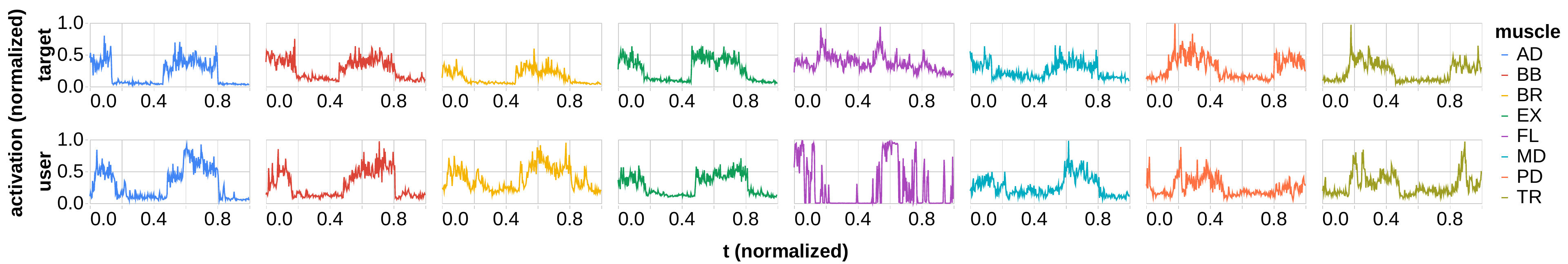}
    \caption{\small When model-prescribed trajectory (\textit{left}), designed to induce target muscle activations (\textit{right}, \textit{top}) is followed by new users, they generate qualitatively and qualitatively similar activations (\textit{right}, \textit{bottom} for exemplar new user P17).
    \vspace{-1em}
    }
    \label{fig:closed-loop:1005}
\end{figure}

\textbf{Trajectory prediction performance results \& ablations.}
Table~\ref{tab:results} aggregates prediction performance of the ``Full'' model across test sets, alongside naive baselines (MLP and LSTM) and ablated models, and Figure~\ref{fig:baseline} illustrates exemplar Full model performance. A complete description of all tested architectures can be found in Appendix~\ref{app:abl}. The Full model reliably achieves $\approx$8.5\% deterministic transformer-decoded error, with flow matching providing a performance boost of $\approx$1--2\%, across all test sets, demonstrating superior generalization performance as compared with all ablations. 
The naive MLP baseline is uniformly less performant than the Full model on all test sets. The LSTM baseline achieves deterministic error comparable to our Full model's $\mathrm{MSE}_\mathrm{FM}$ on new tasks performed by known users and on new tasks performed by new users, but generalizes poorly to new users performing known tasks, and also performs poorly on the training set (a meaningful deficiency, as rehabilitation often consists of patients returning to complete similar tasks). For almost all models and ablations, standard deviation is substantially higher than mean error, and Figure~\ref{fig:baseline} presents exemplar data illustrating the source of this variance in our Full model: portions of the trajectory within the distribution of the test set exhibit substantially higher prediction accuracy than out-of-distribution areas, which flow matching tends to pull toward the training distribution.

\begin{table}[t]
\footnotesize
    \def\arraystretch{1.1}%
    \caption{Trajectory Prediction Performance of Baselines, Ablated Models, and Full Model}
    \label{tab:results}
    \renewcommand{\arraystretch}{1.3}
    \begin{tabular}
        {>{\raggedright\arraybackslash}p{0.09\textwidth} | >{\raggedright\arraybackslash}p{0.13\textwidth} | >{\raggedright\arraybackslash}p{0.04\textwidth} | >{\raggedright\arraybackslash}p{0.04\textwidth} | >{\raggedright\arraybackslash}p{0.04\textwidth} | >{\raggedright\arraybackslash}p{0.09\textwidth} | >{\raggedright\arraybackslash}p{0.09\textwidth} | >{\raggedright\arraybackslash}p{0.09\textwidth} | >{\raggedright\arraybackslash}p{0.11\textwidth}}
        \hline
        Metric & Model & \multicolumn{3}{c|}{Components} & \multicolumn{4}{c}{Error (mean $\pm$ standard deviation)} \\
        & & Traj. Dec. & Act. Dec. & Flow Mat. & \textit{Training Set} & New Tasks & New Users & New Tasks \& Users  \\ \hline
        $\mathrm{MSE_{TD}}$ & Naive MLP      & $\checkmark$ &              &              & $\mathit{0.106} \pm \mathit{0.273}$ & $0.097 \pm 0.233$ & $0.238 \pm 0.462$ & $0.106 \pm 0.204$ \\
        &    LSTM       & $\checkmark$ &              &              & $\mathit{0.155} \pm \mathit{0.308}$ & $0.068 \pm 0.143$ & $0.191 \pm 0.354$ & $0.063 \pm 0.128$ \\
        & Minimal       & $\checkmark$ &              &              & $\mathit{0.016} \pm \mathit{0.109}$ & $0.108 \pm 0.211$ & $0.099 \pm 0.256$ & $0.115 \pm 0.213$ \\
        & + Act. Dec.   & $\checkmark$ &  $\checkmark$  &  & $\bfitnum{0.005} \pm \bfitnum{0.074}$ & $0.117 \pm 0.267$ & $0.088 \pm 0.252$ & $0.110 \pm 0.228$  \\
        & + Flow Mat.   & $\checkmark$ &  & $\checkmark$  & $\mathit{0.069} \pm \mathit{0.216}$ & $0.115 \pm 0.239$ & $0.095 \pm 0.255$ &  $0.106 \pm 0.185$ \\
        &    {Full} & $\checkmark$ & $\checkmark$ & $\checkmark$ & $\mathit{0.009} \pm \mathit{0.110}$ & $0.084 \pm 0.227$ & $0.086 \pm 0.252$ & $0.082 \pm 0.211$  \\
        \hline
        $\mathrm{MSE_{FM}}$ & + Flow Mat.   & $\checkmark$ &              & $\checkmark$ & $\mathit{0.076} \pm \mathit{0.213}$ & $0.114 \pm 0.241$ & $0.098 \pm 0.246$ & $0.106 \pm 0.196$ \\
        & {Full} & $\checkmark$ & $\checkmark$ & $\checkmark$ & $\mathit{0.010} \pm \mathit{0.094}$ & $\mathbf{0.067} \pm \mathbf{0.190}$ & $\mathbf{0.077} \pm \mathbf{0.211}$ & $\mathbf{0.062} \pm \mathbf{0.174}$ \\ \hline
    \end{tabular}
\end{table}

\textbf{Preliminary trajectory prescription performance.} 
To show proof of concept of intended model usage --- closed-loop prescription of trajectories to induce specified muscle activations ---
we selected an exemplar set of muscle activations from an arbitrary system user and task (P07, T08) and generated a new trajectory from those muscle activations using our most performant (Full) model (expected trajectory from 250 samples of the flow-matching model; Figure~\ref{fig:closed-loop:1005}, \textit{Left}). Five participants (P08 and P13 from the original cohort, and P17--P19 new) were then instructed to complete this new trajectory. Participants were instructed to follow the trajectory as closely as possible, and were granted multiple chances to repeat the trajectory to fix their mistakes if requested. We only report on these participants' final (not best) trajectory repetition. Figure~\ref{fig:closed-loop:1005} (\textit{Right}) shows exemplar prescription performance for new participant P17, illustrating the reliable qualitative similarity between target and user-generated muscle activations.\footnote{See Appendix~\ref{app:closed-loop-evidence} for corresponding plots for all users.} Quantitatively, activation error (measured via RMSE) was consistent across all 5 participants, with mean $0.155$ and standard deviation $0.026$.

\textbf{Discussion.}
Our trained model's rapid inference time affirms its suitability for real-time trajectory prescription applications in therapy settings, with solid prediction error for unseen tasks and users; desirable conservative, training-distribution-seeking behavior when encountering out-of-distribution data; and promising proof-of-concept performance inducing target activations through prescribed trajectories.
Moving forward, we aim to
characterize first, the prediction accuracy required to induce therapeutically meaningful muscle activations, and second, model performance in cohorts with specific neuromotor pathologies.

\vspace{-0.5em}
\subsection{User-Specific Fine-Tuning \& Evaluation}\label{sec:finetuning}
\vspace{-0.5em}

In clinical settings, patients 
are likely to exhibit out-of-distribution behavior; while flow matching biases our prescribed trajectories toward those observed in training (which are more likely to be safe, assuming trajectories executed by one user are safe for another), they may not be sufficiently expressive. We present the following method to calibrate our baseline model to new users and remark on the trade-offs between calibration time and model performance. 

\textbf{Fine-tuning procedures.}
To emulate the process of a new user entering a therapy session, performing a calibration task, and engaging with the fine-tuned model, we retrain our baseline model with one or more tasks greedily selected from the new user's data set via an algorithm detailed in Appendix~\ref{app:usrtun}. We evaluate loss $\mathcal{L}$ (as specified in Equation~\ref{eq:loss}), an upper bound on expected trajectory prediction error, for retrained models incorporating as few as 1 and as many as 12 user-specific tasks, on all remaining test tasks, for each test user.

\begin{wrapfigure}{r}{0.5\textwidth}
    \centering
    \vspace{-1em}
    \includegraphics[width=1.0\linewidth]{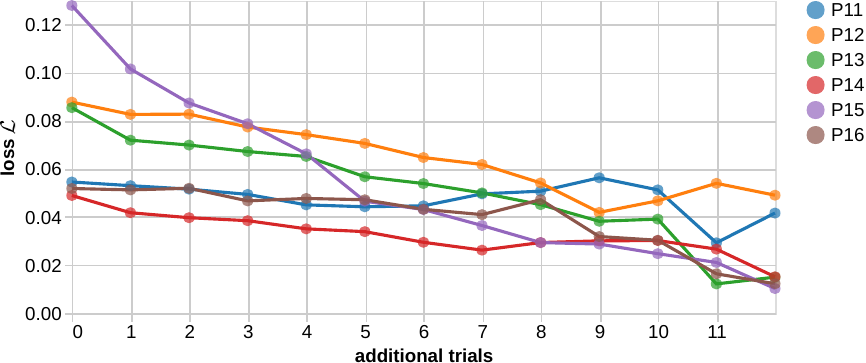}
    \caption{\small Training loss $\mathcal{L}$ upper-bounding trajectory prediction error after re-training using an incrementing number of each user's trials via our greedy re-training algorithm. All users exhibit decreasing error with additional training, characterizing the trade-off between calibration time and prediction accuracy.}
    \label{fig:usrtun}
\end{wrapfigure}

\textbf{Fine-tuning performance \& time requirements.}
Figure~\ref{fig:usrtun} illustrates the loss achieved for each test user after training with incrementing numbers of test tasks. Training performance improved for all users with the incorporation of increasing amounts of user-specific data, with variable effect sizes ranging from $\approx$2--12\%. The time required to train all 12 models for each participant was $\approx$7~min, equally distributed across models, such that each generated model represents $\approx$35~s in training time. This means that the total calibration time, accounting for both additional data collection time and training, can be interpreted as ${n_\mathrm{trials}}_{\mathrm{new}} \times \{60, 20\} + 35$~s (for 60 and 20~s tasks, respectively).

\textbf{Discussion.}
Short training times (a few minutes in the context of a $\approx$30--60~min therapy appointment) and modest-to-substantial performance improvements in our pilot cohort demonstrate the feasibility of meaningfully fine-tuning our model on clinical timelines.

\vspace{-0.5em}
\section{Limitations}
\vspace{-0.5em}

This paper presents the first proof-of-concept usage of our trajectory prescription model. 
Testing with target populations will be needed to establish therapeutic utility. Specific therapy tasks used in training and testing have also not been optimized; we hypothesize that performance could be substantially enhanced by training on more diverse trajectories rather than stereotyped motions along each axis at limited force levels, and determining how to select such trajectories optimally for most efficient training is a rich area for future exploration.
Now that prescription feasibility is established, future work includes model training with larger, pathology-specific cohorts, including combination with a pathology- or user-specific residual model, and consideration of additional factors known to influence neuromotor behavior, such as handedness (5 of our 16 participants were left-handed, but all completed tasks with their right arm).
The behavior of our flow matching algorithm near the edges of the training distribution has also only been examined qualitatively, and its uncertainty estimates should not be interpreted as a safety guarantee; for more robust safety in deployment, our model can be combined with multiple layers of safety guarantees, such as hardware joint/speed limits and control barrier function and/or reachability-based control layers.
In addition, hyperparameter values were largely selected by intuition, and a more systematic hyperparameter search could likely also improve performance.
Lastly, all interactions were isometric, and we aim to expand this model in the future to prescribe dynamic end effector trajectories.

\vspace{-0.5em}
\section{Conclusions}
\vspace{-0.5em}
We have demonstrated a first-of-its-kind activation-optimizing trajectory prescription model shown to generalize to unseen users and tasks, run fast enough for real-time therapy use, and be fine-tuned with brief user-specific calibration. These results support the feasibility of activation-targeted robot-mediated rehabilitation, and we have open-sourced all data and code to facilitate such development.

\clearpage

\acknowledgments{
    The authors thank Ajay Anand and Jono Jenkens for their contributions to the development of the system that enabled data collection for this study. We also thank the reviewers for their careful reading of the manuscript and their constructive, detailed feedback, which led to meaningful improvements, including the live demonstration of our model's closed-loop capabilities and a more rigorous description of the limitations of our model.
}

\bibliography{refs}

\clearpage

\appendix

\section*{Appendices}

\section{Participant Demographics}\label{app:pardem}

Table~\ref{tab:participants} contains all participant demographic data. Note that all participants completed tasks with their right arm regardless of handedness.

\begin{table}[h]
\def\arraystretch{1.3}%
\centering
\caption{Participant Demographics}
\begin{tabular}{lrccl}
\hline
\textbf{ID} & \textbf{Age} & \textbf{Gender} & \textbf{Handedness} \\
\hline
P01 & 21 & M & Left \\
P02 & 24 & M & Right \\
P03 & 24 & M & Right \\
P04 & 19 & M & Right \\
P05 & 27 & M & Left \\
P06 & 47 & F & Right \\
P07 & 31 & F & Right \\
P08 & 33 & F & Right \\
P09 & 34 & M & Right \\
P10 & 30 & F & Right \\
P11 & 37 & F & Right \\
P12 & 23 & M & Left \\
P13 & 25 & M & Right \\
P14 & 29 & M & Right \\
P15 & 23 & M & Left \\
P16 & 28 & M & Left \\
P17 & 23 & F & Left \\
P18 & 28 & M & Left \\
P19 & 29 & M & Right \\
\hline
\end{tabular}
\label{tab:participants}
\end{table}

\section{Protocol Deviations}\label{app:prodev}
Our first participant, P01, completed tasks at lower and higher force values ($15 \rightarrow 5$, $36 \rightarrow 20$, $30 \rightarrow 50$, $72 \rightarrow 120$  N) which were determined to be artificially low and unsustainably high, respectively. These protocol deviations reflect improvements made during system development: P01 provided critical feedback on required force levels, which were adjusted for subsequent participants to provide a more comfortable user experience. Similarly, T14 and T15 were added after the first two participants to develop a more robust test set, such that P01 and P02 did not complete them. We elected not to exclude these early-protocol data from model training and evaluation, granting our model the widest distribution possible.

\section{Locked \& Unlocked Task Motivation \& Performance}\label{app:lock}
\begin{figure}[H]
    \centering
    \includegraphics[width=1.0\linewidth]{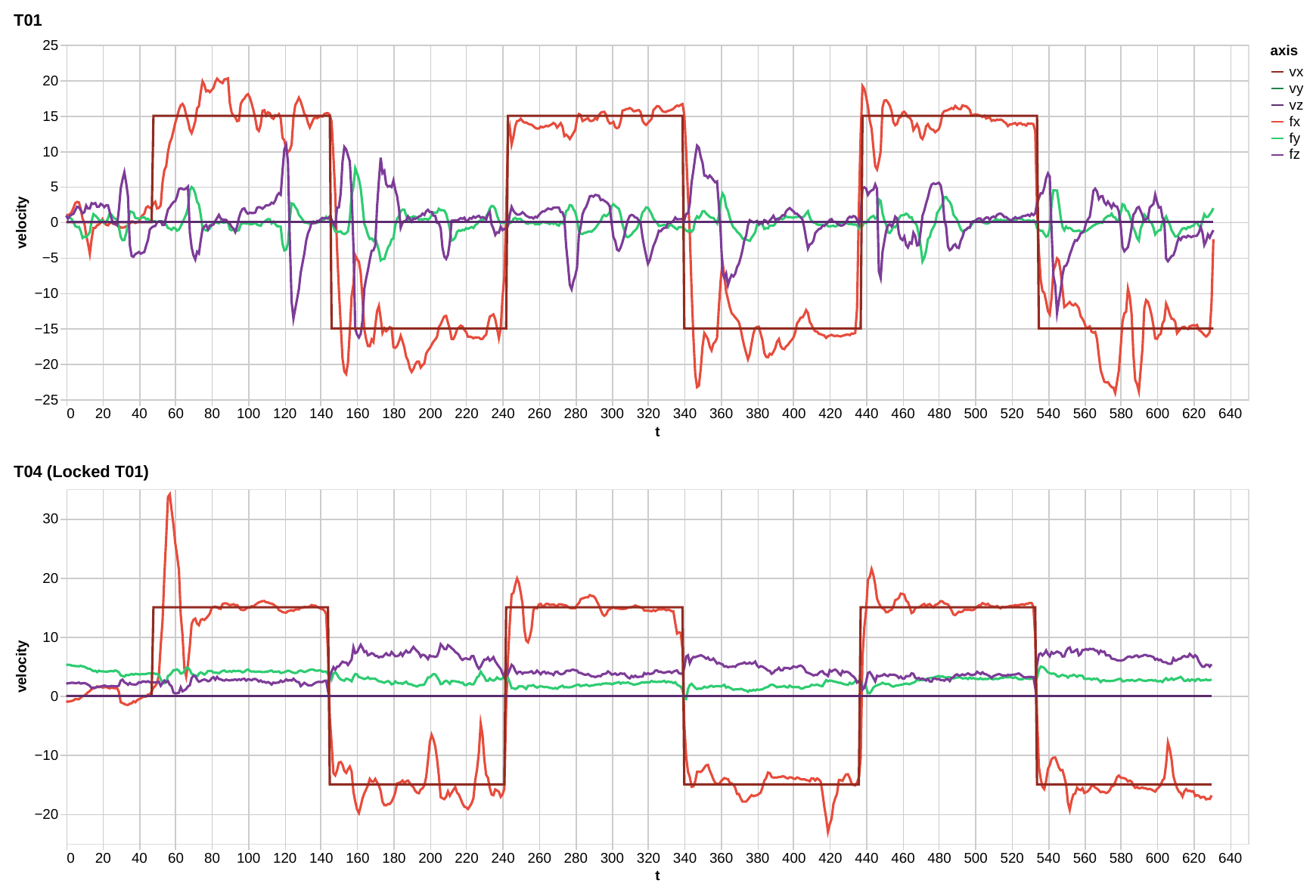}
    \caption{Example of user force generation behavior $\mathbf{f}$ about target $\mathbf{v}$ on unlocked $x$-axis task T01 (\textit{top}) and its locked counterpart T02 (\textit{bottom}). Due to the challenge of maintaining appropriate values in all axes during unlocked tasks, users frequently correct one axis at a time, resulting in oscillating signals. On locked tasks, users' signals are smoother, but have bias in unspecified dimensions that varies from user to user.
    }
    \label{fig:lockunlock}
\end{figure}

Each task was undertaken in both ``unlocked'' and ``locked'' versions, where users' force inputs were not and were constrained to the task's relevant axis, respectively. As illustrated in Figure~\ref{fig:lockunlock}, these tasks resulted in qualitatively different behavior along both prescribed and non-prescribed axes: in unlocked tasks, users oscillated about the target values in all 3 axes, while for locked tasks, users' motions were smoother, but maintained an offset in non-prescribed axes that differed across users. Both locked and unlocked task specifications are used in therapy, and with no consensus on which may perform most effectively, we elect to include both.
A comprehensive examination of model performance across both conditions is beyond the scope of this work, but in practice, the model performed with comparable overall error on both types of tasks.

\section{Task Trajectories}\label{app:traj}
\begin{figure}[H]
    \centering
    \includegraphics[width=1.0\linewidth]{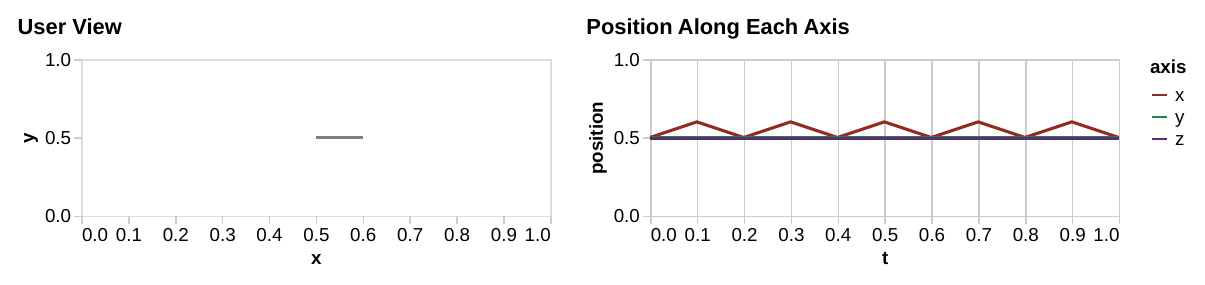}
    \caption{T01 and T04}
    \label{fig:trajectories:14}
\end{figure}

\begin{figure}[H]
    \centering
    \includegraphics[width=1.0\linewidth]{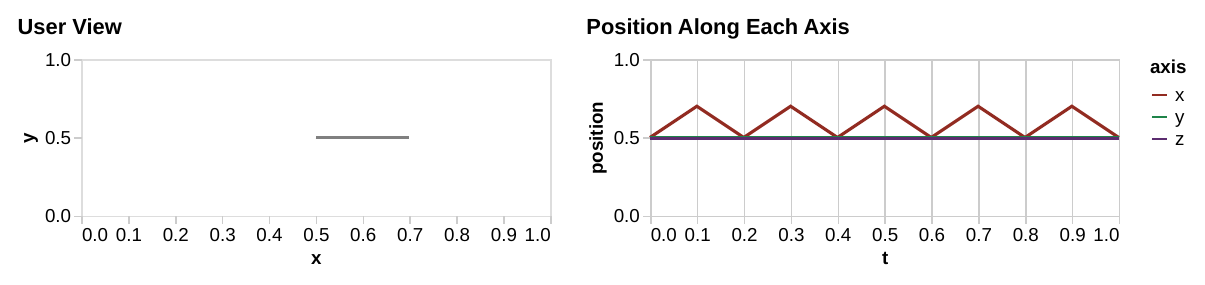}
    \caption{T07 and T10}
    \label{fig:trajectories:710}
\end{figure}

\begin{figure}[H]
    \centering
    \includegraphics[width=1.0\linewidth]{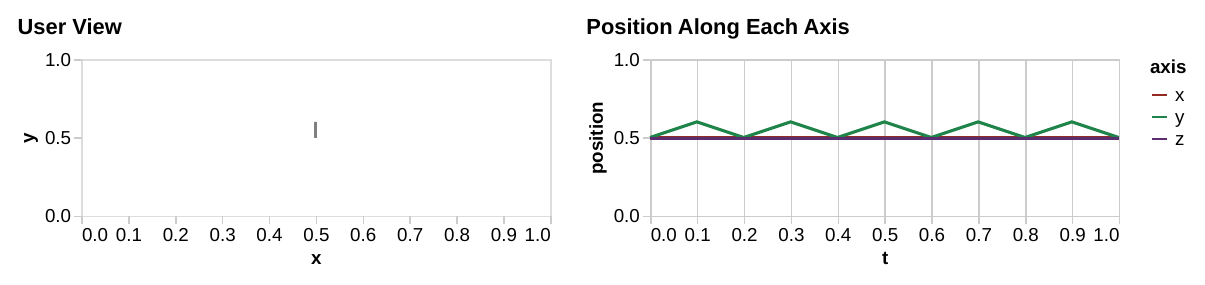}
    \caption{T02 and T05}
    \label{fig:trajectories:25}
\end{figure}

\begin{figure}[H]
    \centering
    \includegraphics[width=1.0\linewidth]{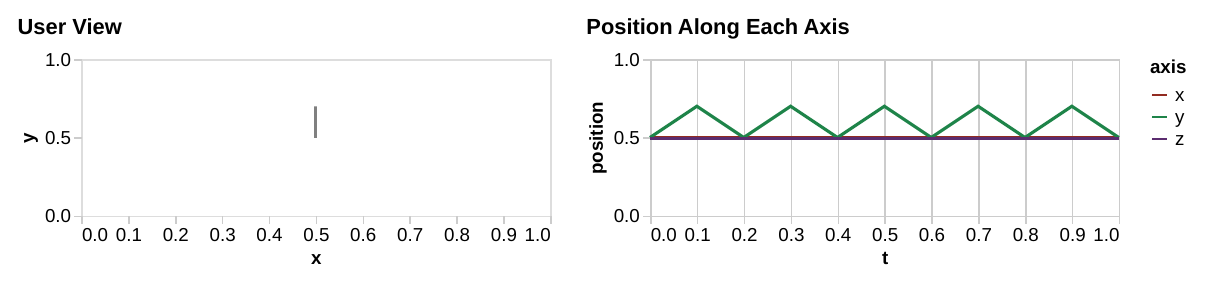}
    \caption{T08 and T11}
    \label{fig:trajectories:811}
\end{figure}

\begin{figure}[H]
    \centering
    \includegraphics[width=1.0\linewidth]{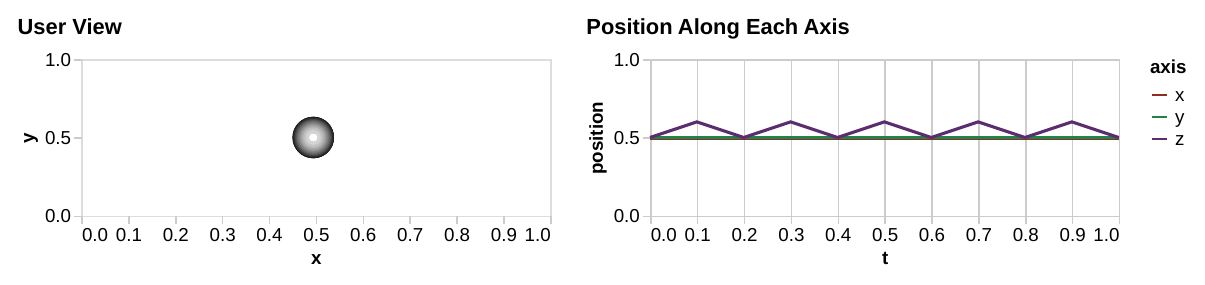}
    \caption{T03 and T06}
    \label{fig:trajectories:36}
\end{figure}

\begin{figure}[H]
    \centering
    \includegraphics[width=1.0\linewidth]{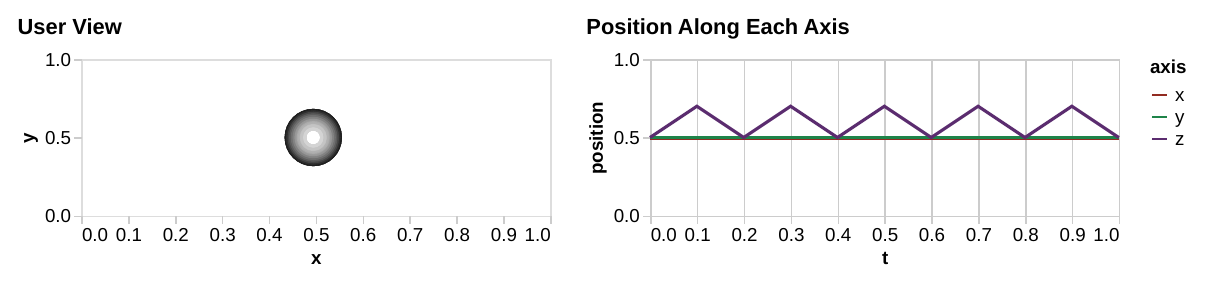}
    \caption{T09 and T12}
    \label{fig:trajectories:912}
\end{figure}

\begin{figure}[H]
    \centering
    \includegraphics[width=1.0\linewidth]{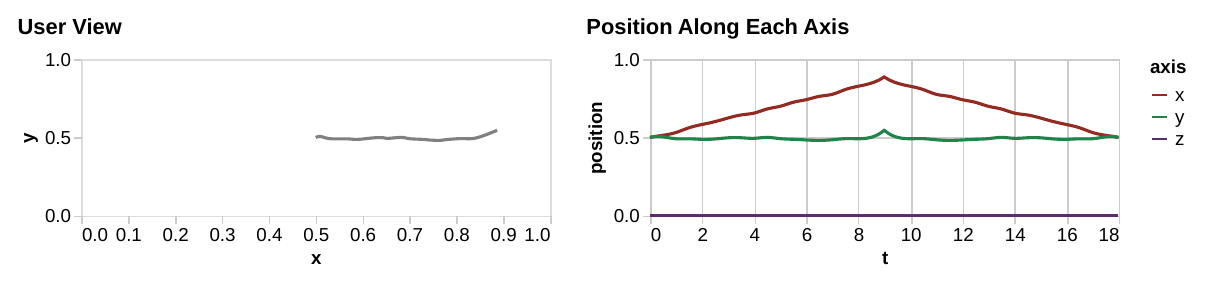}
    \caption{T14 (exemplar)}
    \label{fig:trajectories:stochx}
\end{figure}

\begin{figure}[H]
    \centering
    \includegraphics[width=1.0\linewidth]{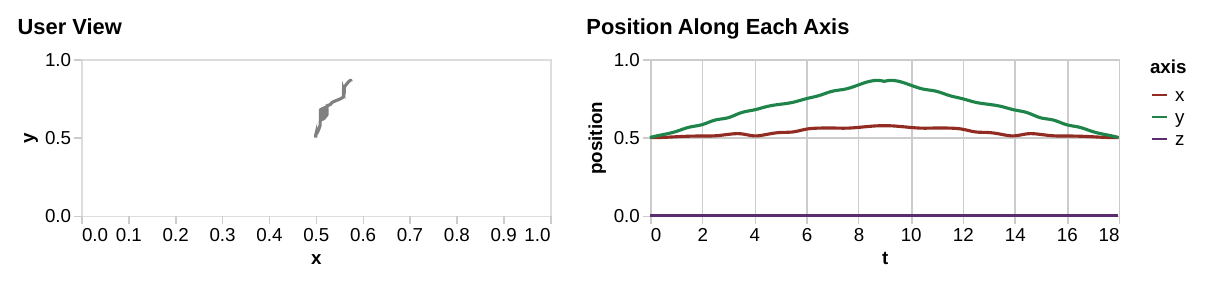}
    \caption{T15 (exemplar)}
    \label{fig:trajectories:stochy}
\end{figure}

Tasks T01--T12 (the training set), illustrated in Figures~\ref{fig:trajectories:14}--\ref{fig:trajectories:912}, consisted of a dominant axis that users were asked to follow. Users were shown the trajectory that their target would follow, a target ball traversing that trajectory at the prescribed pace, and an avatar ball whose velocity they controlled via force exertion. ``Unlocked'' and ``locked'' versions of the tasks shared trajectories; for locked versions, the avatar ball only deviated in appearance along the prescribed axis. For tasks along the $z$-axis, target $z$-coordinate position was indicated by increasing and diminishing size of the target ball (shown in Figures~\ref{fig:trajectories:36}--\ref{fig:trajectories:912}). Tasks T14--T15 (the test set), illustrated in Figures~\ref{fig:trajectories:stochx}--\ref{fig:trajectories:stochy}, consisted of stochastically generated trajectories along the $x$ and $y$ axes. These trajectories were generated via a Brownian motion diffusion process that was then smoothed by a cubic spline. Trajectories were generated at collection time such that they varied for each participant, and were never seen in the training set.

\section{sEMG Processing Protocol}\label{app:emgproc}
Raw 8-channel sEMG were processed by a butterworth band-pass filter (20--500~Hz), a second-order iir notch filter (to remove power line frequencies), rectified, and smoothed by RMS with window size 200. Processed sEMG were resampled or aligned to 1000~Hz to match the sample rate of the captured forces.

\newpage

\section{Hyperparameters}\label{app:hyper}
Table~\ref{tab:hyperparameters} contains hyperparameter values used during training. These values were selected empirically; a systematic hyperparameter search could likely improve model performance.

\begin{table}[h]
\centering
\caption{Hyperparameters}
\label{tab:hyperparameters}
\renewcommand{\arraystretch}{1.1}
\begin{tabular}{llr}
\toprule
\textbf{Component} & \textbf{Hyperparameter} & \textbf{Value} \\
\midrule
\multirow{11}{*}{Transformer Encoder/Decoder}
& Model width & 128   \\
& Attention heads & 4     \\
& Feedforward width & 512 \\
& Encoder layers  & 3     \\
& Decoder layers  & 3     \\
& Latent dimension & 64    \\
& Dropout                                 & 0.1   \\
& Activation loss weight ($\lambda_1$)          & 1.0   \\
& Velocity Trajectory loss weight ($\lambda_2$)          & 2.0   \\
& Flow Matching loss weight ($\lambda_3$) & 1.0  \\
& Learning rate                           & $10^{-3}$ \\
\midrule
\multirow{3}{*}{Flow Matching}
& Latent dimension ($d_z$)               & 64    \\
& Vector field hidden width              & 256   \\
& ODE solver steps                        & 20    \\
\midrule
\multirow{6}{*}{LSTM (baseline)}
& Hidden width                            & 128   \\
& Encoder layers                          & 3     \\
& Decoder layers                          & 3     \\
& Latent dimension                        & 32    \\
& Dropout                                 & 0.1   \\
& Learning rate                           & $10^{-3}$ \\
\midrule
\multirow{7}{*}{Naive MLP (baseline)}
& Hidden width                            & 128   \\
& Encoder layers                          & 3     \\
& Decoder layers                          & 3     \\
& Latent dimension                        & 32    \\
& Activation function                     & ReLU  \\
& Dropout                                 & 0.1   \\
& Learning rate                           & $10^{-3}$ \\
\bottomrule
\end{tabular}
\end{table}

\section{Baseline \& Ablations Details}\label{app:abl}
Ablations to the model's architecture were carried out to validate if the complexity of our model was justified, and our model was compared against two naive baselines. The following models were tested:
\begin{itemize}[leftmargin=1.2em]
    \item {Naive MLP}: Encoder-Decoder style MLP that maps activation window $\mathbf{a}$ directly to target trajectory $\mathbf{\hat v}_\mathrm{target}$.
    \item {LSTM}: Long short-term memory recurrent network mapping activation window $\mathbf{a}$ directly to target trajectory $\mathbf{\hat v}_\mathrm{target}$.
    \item {Minimal}: Target trajectory reconstruction only (Figure~\ref{fig:arch} \textit{blue} and \textit{green} flows only).
    \item {Minimal + Act. Dec.}: Minimal model plus activation decoder (Figure~\ref{fig:arch} \textit{blue}, \textit{green}, and \textit{turquoise} flows only).
    \item {Minimal + Flow Mat.}: Minimal model plus flow matching (Figure~\ref{fig:arch} \textit{blue}, \textit{green}, and \textit{orange} flows only).
    \item {Full}: Full model (Figure~\ref{fig:arch} all).
\end{itemize}

\section{Trajectory Prescription Performance Plots}\label{app:closed-loop-evidence}

\begin{figure}[H]
    \centering
    \includegraphics[width=1.0\linewidth]{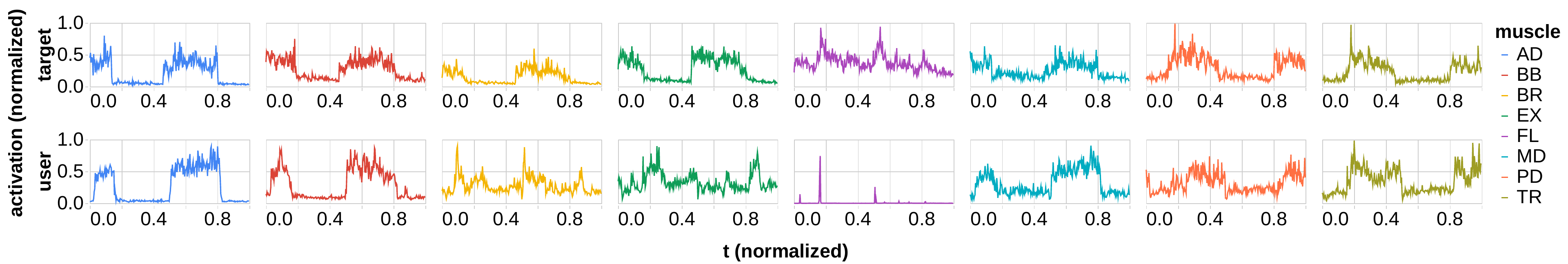}
    \caption{Target (\textit{top}) and user-generated (\textit{bottom}) muscle activations for participant P08.}
    \label{fig:closed-loop:1000}
\end{figure}

\begin{figure}[H]
    \centering
    \includegraphics[width=1.0\linewidth]{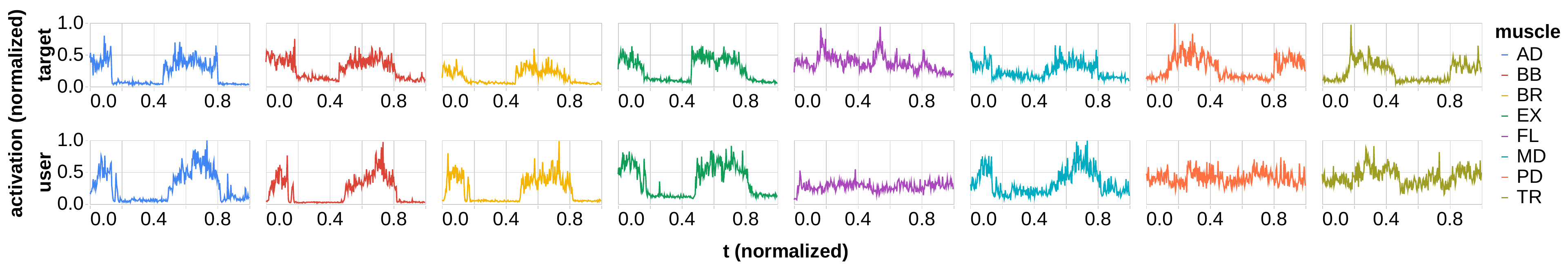}
    \caption{Target (\textit{top}) and user-generated (\textit{bottom}) muscle activations for participant P13.}
    \label{fig:closed-loop:1009}
\end{figure}

\begin{figure}[H]
    \centering
    \includegraphics[width=1.0\linewidth]{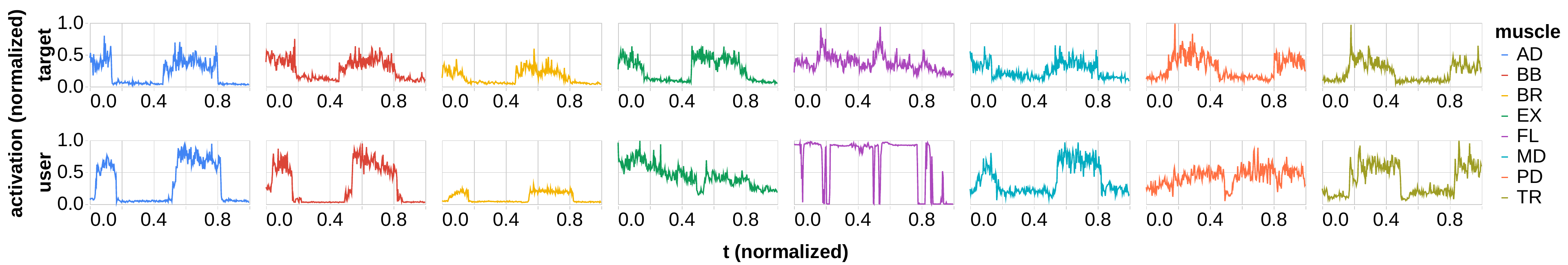}
    \caption{Target (\textit{top}) and user-generated (\textit{bottom}) muscle activations for participant P18.}
    \label{fig:closed-loop:1002}
\end{figure}

\begin{figure}[H]
    \centering
    \includegraphics[width=1.0\linewidth]{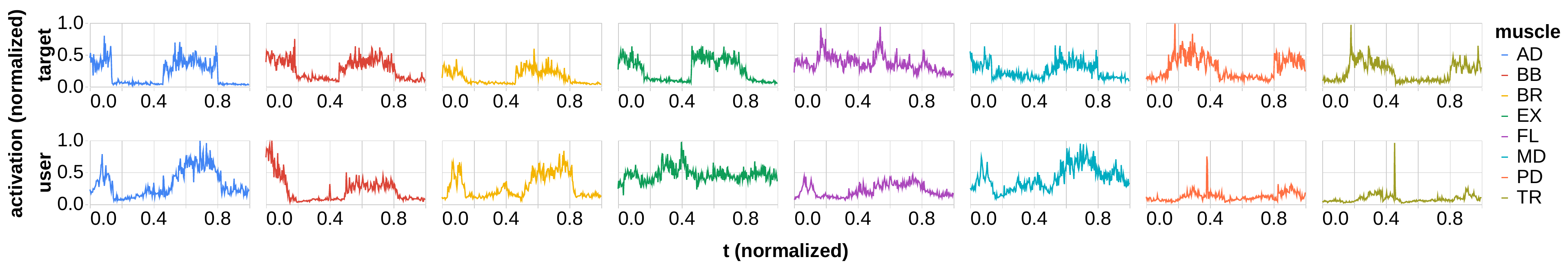}
    \caption{Target (\textit{top}) and user-generated (\textit{bottom}) muscle activations for participant P19.}
    \label{fig:closed-loop:1006}
\end{figure}

\newpage

\section{User-Specific Fine-Tuning Algorithm}\label{app:usrtun}

To identify which tasks were most informative for fine-tuning on unseen users, the greedy algorithm detailed in Algorithm~\ref{alg:gft} was run on user data whose tasks were withheld from the training data. For each held-out user, we loaded all of their task data, loaded the trained weights into our model, trained on each task individually, tested on each task individually, and added the task with the best test loss to a list. This process was repeated until a single task remained, which we retained for our final test set. Empirically, we did not observe substantial difference in performance based on which tasks were selected for training --- more tasks improved performance regardless --- but our small number of tested users (6) means that this remains a preliminary observation. 

\algnewcommand\algorithmicinput{\textbf{Input:}}
\algnewcommand\algorithmicoutput{\textbf{Output:}}
\algnewcommand\Input{\item[\algorithmicinput]}
\algnewcommand\Output{\item[\algorithmicoutput]}

\begin{algorithm}
\caption{GreedyFineTune}
    \begin{algorithmic}[1]
        \Input participant $id$
        \Output ordered sets $\mathcal{T}^*, \mathcal{L}$
        \State $\mathcal{T}^* \leftarrow [\,]$ \Comment{Best task execution order for fine-tuning}
        \State $\mathcal{L} \leftarrow [\,]$ \Comment{Best test loss after each task completion}
        \State $\mathcal{T} \leftarrow \{1, 2, \ldots, 15\}_{id}$
        \While{$|\mathcal{T}^*| < |\mathcal{T}| - 1$}
            \State $S \leftarrow \emptyset$ \Comment{Candidates for best data to use for fine-tuning}
            \For{each task $t \in \mathcal{T}$}
                \State $\mathcal{F} \leftarrow \textsc{LoadPreTrainedModel}(\text{})$ \Comment{Load baseline model}
                \If{$t \in \mathcal{T}^*$}
                    \State \textbf{continue}
                \EndIf
                \State $\mathcal{T}_{\text{train}} \leftarrow \mathcal{T}^* \cup \{t\}$
                \State $\mathcal{T}_{\text{test}} \leftarrow \mathcal{T} \setminus \mathcal{T}_{\text{train}}$
                \State $\textsc{TrainModel}(\mathcal{T}_{\text{train}},\ \mathcal{F})$ \Comment{Re-train model using the new training set}
                \State $\ell \leftarrow \textsc{TestModel}(\mathcal{T}_{\text{test}},\ \mathcal{F})$ \Comment{Re-test model using remaining test set}
                \State $S \leftarrow S \cup \{(t,\ \ell)\}$
            \EndFor
            \State $(t^*, \ell^*) \leftarrow \arg\underset{\ell}{\min}\ S$
            \State $\mathcal{T^*} \leftarrow \mathcal{T^*} \cup [t^*]$
            \State $\mathcal{L} \leftarrow \mathcal{L} \cup [\ell^*]$
        \EndWhile
        \State $\mathcal{T}_{\text{rem}} \leftarrow \mathcal{T} \setminus \mathcal{T}^*$
        \State \Return $\mathcal{T}^* \cup \mathcal{T}_{\text{rem}}$, $\mathcal{L}$
    \end{algorithmic}
    \label{alg:gft}
\end{algorithm}

\end{document}